\documentclass[a4paper,conference]{IEEEtran}

\ifCLASSINFOpdf
  \usepackage[pdftex]{graphicx}
  \usepackage{hyperref}
  \usepackage{times}
  \usepackage{epsfig}
  \usepackage{amsmath}
  \usepackage{amssymb}
\else
\fi

\begin{document}

\title{Unsupervised Multiple Person Tracking\\ using AutoEncoder-Based Lifted Multicuts}

\author{\IEEEauthorblockN{Kalun Ho}
\IEEEauthorblockA{Fraunhofer Center Machine Learning\\
Fraunhofer ITWM\\
Email: kalun.ho@itwm.franhofer.de }
\and
\IEEEauthorblockN{Janis Keuper}
\IEEEauthorblockA{Institute for Machine Learning \\
and Analytics (IMLA)\\
Offenburg University, Germany\\
Email: keuper@imla.ai}
\and
\IEEEauthorblockN{Margret Keuper}
\IEEEauthorblockA{University of Mannheim, Germany\\
Email: keuper@uni-mannheim.de}}

\maketitle

\begin{abstract}
Multiple Object Tracking (MOT) is a long-standing task in computer vision. Current approaches based on the tracking by detection paradigm either require some sort of domain knowledge or supervision to associate data correctly into tracks. In this work, we present an unsupervised multiple object tracking approach based on visual features and minimum cost lifted multicuts. Our method is based on straight-forward spatio-temporal cues that can be extracted from neighboring frames in an image sequences without superivison. Clustering based on these cues enables us to learn the required appearance invariances for the tracking task at hand and train an autoencoder to generate suitable latent representation. Thus, the resulting latent representations can serve as robust appearance cues for tracking even over large temporal distances where no reliable spatio-temporal features could be extracted. We show that, despite being trained without using the provided annotations, our model provides competitive results on the challenging MOT Benchmark for pedestrian tracking.
\end{abstract}

\section{Introduction}

The objective of multiple object tracking is to find a trajectory for each individual object of interest in a given input video. 
Specific interest has been devoted to the specific task of multiple person tracking~\cite{zamir2012gmcp,henschel2017improvements,tang2016multi,tang2017multiple,luo2014multiple}. 
Most successful approaches follow the \textit{Tracking-By-Detection} paradigm.
First, a state-of-the-art detector is used in order to retrieve the position of each object (pedestrian) within each frame. 
Secondly, the output detections of same persons across video frames are associated over space and time in order to form unique trajectories. 
This matching of detections can also be considered as a clustering problem with an unknown number of clusters. 
Since objects might get occluded during the video sequence or the detector might simply fail on some examples, successful approaches are usually based not solely on spatial but also on appearance cues, learned on annotated data using, for example, Siamese networks for person re-identification~\cite{tang2017multiple}.

\begin{figure}[t]
	\begin{center}
		\includegraphics[width=0.85\linewidth]{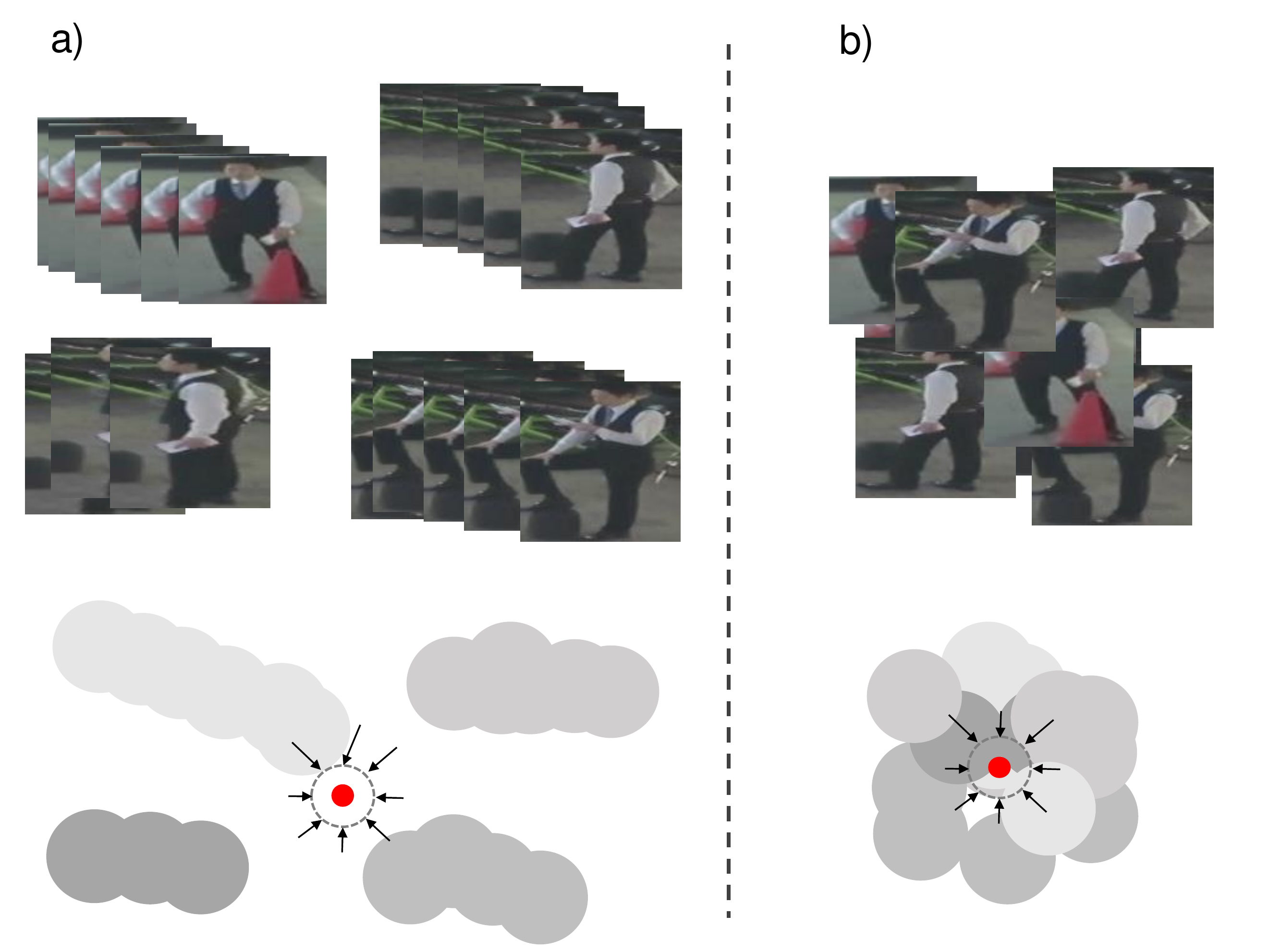}
	\end{center}
	\caption{Visualization of the proposed idea: a) Appearance centroid from spatio-temporal information. b) The appearance space is trained to minimize the distance between centroid and associated detections.}
	\label{fig:teaser}
\end{figure}

\textbf{Motivation.}
Supervised approaches to person re-identification require large amounts of sequence specific data in order to achieve good performance, which is why multiple object tracking benchmarks such as MOT~\cite{MOT16} provide, for every test sequence, a training sequence recorded in a sufficiently similar setting. 
Yet, it has been shown for example in \cite{yoon2018online, feng2019multi} that incorporating additional training data to the model training on the multiple object tracking benchmark~\cite{MOT16} is key to improving the overall tracking performance. 

Thus, publicly available, annotated training data currently seems not to be sufficient for training reliable person re-identification networks. 
Yet, recording such data in a setting close to a final test scenario usually comes at a low price. 
Thus, the need for methods with a low amount of supervision becomes obvious and motivates us to propose a multiple object tracking method based on self-supervision. 
Yet, learning suitable object appearance metrics in an unsupervised way is challenging since, compared to classical clustering problems, visual features of the same person may change over time due to pose and viewpoint changes.
Other issues such as frequent and long range occlusion or background noises makes pedestrian tracking even more challenging.

Recently, Deep Neural Network (DNN) has gained significant role in research and industry. DNN enables "learning from data", therefore it is often used when a significant large amount of data is available where the goal is to find patterns or features in it. By extracting these features, one can build a model to correctly classify unseen data if the correct labels are provided. 
This is called supervised learning. 
On the other hand, unsupervised approaches group similar data together into clusters.
This is often more challenging as the model has to associate the correct data together without any ground truth (target value). 

In this paper, we propose an approach for learning appearance features for multiple object tracking without utilizing human annotations of the data. 
Our approach is, on the one hand, based on the observation that, from an image sequence, many data associations can be made reliably from pure spatio-temporal cues such as the intersection over union (IoU) of bounding boxes within one frame or between neighboring frames. 
Resulting tracklets, on the other hand, carry important information about the variation of an objects appearance over time for example by changes of the pose or viewpoint. 
In our model, we cluster the initial data based on simple spatial cues using the recently successful minimum cost multicut approach \cite{tang2016multi}. 
Resulting clustering information is injected into a convolutional auto-encoder to enforce detections with the same, spatio-temporally determined label to be close to one-another in the latent space (see Fig.\ref{fig:teaser}). 
Thus, the resulting latent data representation encode not only the pure object appearance but also the expected appearance variations within one object ID. 
Distances between such latent representations can serve to re-identify objects even after long temporal distances, where no reliable spatio-temporal cues could be extracted. 
We use the resulting information in the minimum cost lifted multicut framework, similar to the formulation of Tang~\cite{tang2017multiple}, whose method is based on Siamese networks trained in a fully supervised way.

To summarize, our contributions are:
\begin{itemize}
    \item We present an approach for multiple object tracking, including long range connections between objects, which is completely supervision-free in the sense that no human annotations of the data are employed.
	\item We propose to inject spatio-temporally derived information into convolutional auto-encoders in order to produce a suitable data embedding space for multiple object tracking.
	\item  We evaluate our approach on the challenging MOT17 benchmark and show competitive results without using actual supervision. 
\end{itemize}

The rest of the paper is structured as follows. Section \ref{sec:related} discusses the related work on multiple object tracking. Our unsupervised approach on multiple object tracking is explained in Section \ref{subsec:method}. In Section \ref{sec:results}, we show the tracking performance of our proposed method in the MOT Benchmark ~\cite{MOT16}\footnote{\url{https://motchallenge.net/data/MOT17/}}. 
A conclusion is summarized in Section \ref{sec:conclusion}.

\section{Related Work}
\label{sec:related}
In Multiple Object Tracking, the most common approach is \textit{Tracking-by-Detection}. 
The objective is to associate detections of invidivual persons, which may have spatio or temporal changes in the video. 
Thus re-identification over a long range remains a challenging task.
Multiple object tracking by linking bounding box detections (\emph{tracking by detection}) was studied, e.g., in \cite{Pirsiavash:2011:GOG,Andriyenko2012CVPR,Huang:2008:ROT,AndrilukaCVPR2010,FragkiadakiECCV12,Zamir:2012:GMC,Henschel:2014:EMP,tang14ijcv,Henschel:2014:EMP,DBLP:journals/corr/HenschelLCR17}. 
These works solve the combinatorial problem of linking detection proposals over time via different formulations e.g. via integer linear programming \cite{Shitrit:2011:TMP,wang-et-al-2014}, MAP estimation~\cite{Pirsiavash:2011:GOG}, CRFs \cite{10.1007/978-3-319-16817-3_29}, continuous optimization \cite{Andriyenko2012CVPR} or dominant sets \cite{7503631}. 
In such approaches, the pre-grouping of detections into tracklets or non-maximum suppression are commonly used to reduce the computational costs \cite{Huang:2008:ROT,WojekECCV10,AndrilukaCVPR2010,FragkiadakiECCV12,Zamir:2012:GMC,WojekPAMI2013,Henschel:2014:EMP,tang14ijcv}.
For example Zamir~et~al.~\cite{Zamir:2012:GMC} use generalized minimum clique graphs to generate tracklets as well as the final object trajectories.
Non-maximum suppression also plays a crucial role in disjoint path formulations, such as \cite{networkflow1,networkflow2,Chari2015OnPC}.

In the work of Tang et al. \cite{tang2016multi}, local pairwise features based on DeepMatching are used to solve a multi-cut problem. 
The affinity measure is invariant to camera motion and thus makes it reliable for short term occlusions. 
An extension of this work is found in \cite{tang2017multiple}, where additinal long range information is included. 
By introducing a lifted edge in the graph, improvement of person re-identification is achieved.
In \cite{keuper2018motion}, low-level point trajectories and the detections are combined to solve a co-clustering problem jointly, where dependencies are established between the low-level points and the detections. 
Henschel et al. \cite{henschel2018fusion} solves the multiple object tracking problem by incorporating addtional head detecion to the full body detection while in \cite{henschel2019multiple}, they use a body and joint detector to improve the quality of the provided noisy detections from the benchmark.
Other works that treat Multiple object tracking as a graph-based problem can be found in \cite{henschel2017improvements}, \cite{keuper2015efficient}, \cite{keuper2016multi}, \cite{kumar2014multiple} and \cite{zamir2012gmcp}.
In contrast, \cite{ma2018trajectory} introduces a tracklet-to-tracklet method based a combination of Deep Neural Networks, called \textit{Deep Siamese Bi-GRU}. 
The visual appearance of detections are extracted with CNNs and RNNs in order to generate a tracklet of individuals.
These tracklets are then split and reconnected in a way that occluded persons are correctly re-identified. 
The framework uses spatial and temporal information from the detector to associate the tracklet. 
The approach in~\cite{bergmann2019tracking} exploits the bounding box information by learning from detectors first and combined with a re-identification model trained on a siamese network. 
While the state of the art approaches in MOT17 Challenge are all based on supervised learning \cite{henschel2018fusion, keuper2018motion, kim2015multiple, 8533372, chen2017enhancing}, there are similar works in \cite{li2018unsupervised, lv2018unsupervised}, which attempt to solve person re-identification (ReID) problems in an unsupervised manner.

\section{AutoEncoder-Based Multicut Approach}
\label{subsec:method}
\begin{figure*}[t]
	\begin{center}
		\includegraphics[width=1.0\linewidth]{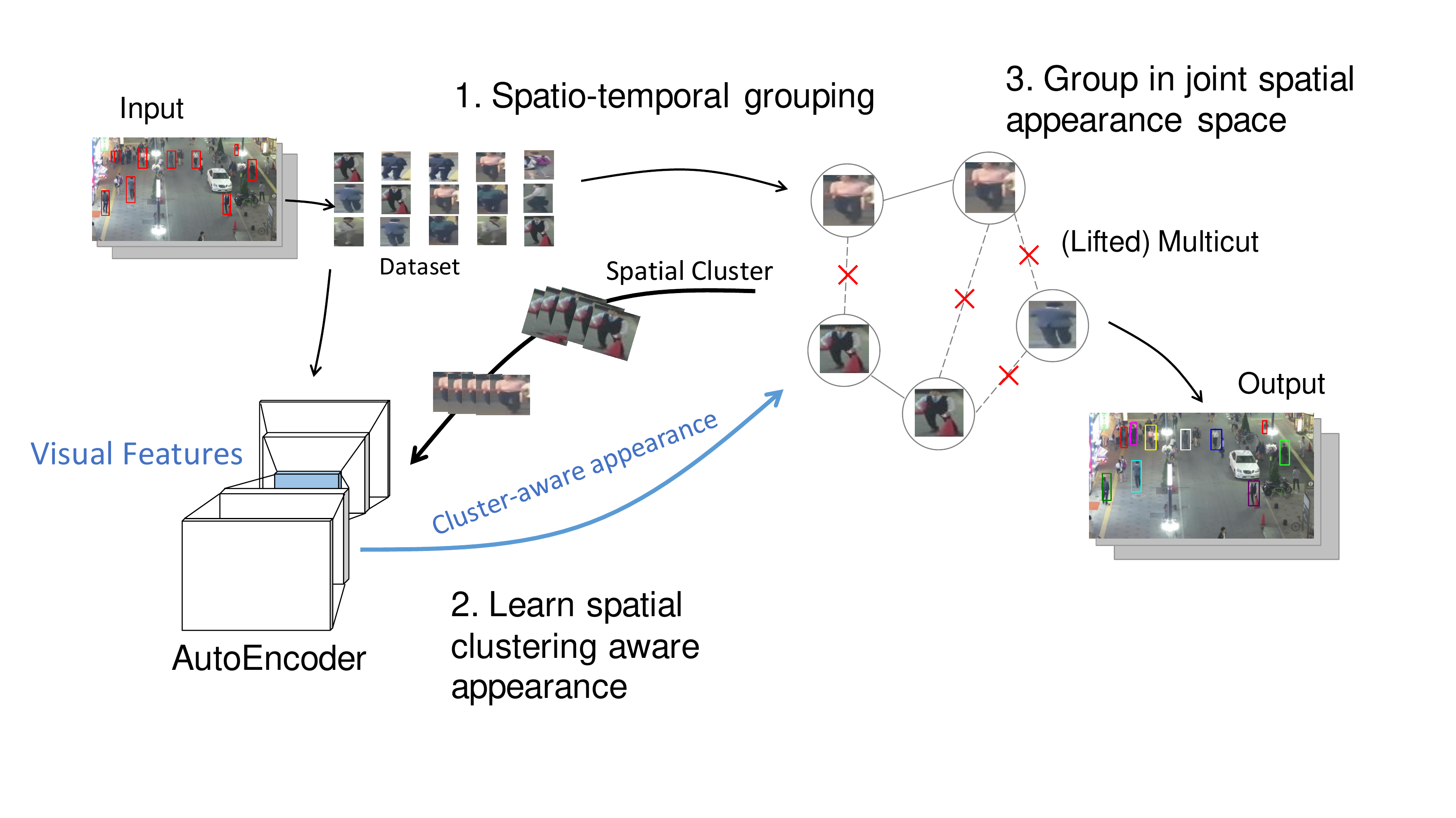}
	\end{center}
	\caption{Summary of our approach in three steps. 1. Initial multicut clustering to obtain weak cluster labels (tracklets), 2. Learn visual features by an AutoEncoder, 3. Extract affinity information from AutoEncoder and compute final clustering (tracking) using lifted multicuts.}
	\label{fig:figure0}
\end{figure*}

The proposed approach is based on the idea to learn, from simple spatial data associations between object detections in image sequences, which appearance variations are to be expected within one object for the task of multiple object tracking. An overview of our workflow implementing this idea is given in Fig.~\ref{fig:figure0}.

First, object detection bounding boxes are extracted along with their spatial information such that spatial correspondences between detections in neighboring frames can be computed.
Based on these simple spatial associations, detections can be grouped into tracklets using clustering approaches such as correlation clustering, also referred to as minimum cost multicuts~\cite{demaine-2006}.

Secondly, a convolutional auto-encoder is trained to learn the visual features of detections. 
The objective is to learn a latent space representation which can serve to match the same object in different video frames. 
Thus, the information about spatial cluster labels from the first step is used as the centroid of latent features. 
Distances between latent representations of data samples and their centroids are minimized in the convolutional AutoEncoder using a clustering loss. 

Lastly, the data are transformed into the latent space of the trained auto-encoder to extract pairwise appearance distances which are expected to encode the desired invariances. Such pairwise appearance distances are used to not only provide additional grouping information between nearby detections, but also for detections with long temporal distance. The final detection grouping is computed using minimum cost lifted multicuts~\cite{keupericcv}.

This section is divided into three subsections: 
Section \ref{subsec:multicut} describes the minimum cost (lifted) multicut approach employed for obtaining the initial spatial cluster labels (e.g. tracklets), as well as for the generation of the final tracking result. Section \ref{subsec:autoencoder} describes the feature learning process using a convolutional AutoEncoder and cluster labels, and section~\ref{subsec:affinity} describes the computation of the joint spatial and appearance metrics used in the final data association step within the minimum cost lifted multicut framework.

\subsection{Multicut Formulation}
\label{subsec:multicut}

We follow Tang~\cite{tang2017multiple} and phrase the multiple target tracking problem as a graph partitioning problem, more concretely, as a minimum cost (lifted) multicut problem. This formulation can serve as well for an initial tracklet generation process, which will help us to inject cues learned from spatial information into the appearance features, as it can be used to generate the final tracking result using short- and long-range information between object detections. 

\subsubsection{Minimum Cost Multicut Problem}
We assume, we are given an undirected graph {\it $G = (V, E)$}, where nodes $v\in V$ represent object detections and edges $e\in E$ encode their respective spatio-temporal connectivity. Additionally, we are given real valued costs {\it $c: E \rightarrow \mathbb{R}$} defined on all edges. Our goal is to determine \emph{edge} labels {\it $y: E \rightarrow \{0, 1\}$} defining a graph decomposition such that every partition of the graph corresponds to exactly one object track (or tracklet). To infer such an edge labeling, we can solve instances of the minimum cost multicut problem with respect to the graph G and costs c, defined as follows \cite{chopra-1993,demaine-2006}:

\begin{align}
\min\limits_{y \in \{0, 1\}^E}
\sum\limits_{e \in E} c_e y_e
\label{eq:mc}
\end{align}

\begin{align}
s.t. \quad \forall C \in cycles(G) \quad \forall e \in C : y_e \leq \sum\limits_{e^\prime \in C\backslash\{e\}} y_{e^\prime}
\label{eq:cycle}
\end{align}
Here, the objective is simple to cut those edges with negative costs $c_e$ such that the resulting cut is a decomposition of the graph. This condition is formalized by the cycle inequalities in Eq.~\eqref{eq:cycle}, which make sure that, for every cycle in $G$, if one of its edges is cut, so is at least one other. Thus, no two nodes can remain connected via some path of the graph if an edge is cut between them along any other path. 

In \cite{chopra-1993}, it was shown to be sufficient to enforce Eq.~\eqref{eq:cycle} on all \emph{chordless} cycles, i.e. all cycles.

Typically, if cut probabilities between pairs of nodes are available, the costs are computed using the \emph{logit} function $\text{logit}(p) = \log\frac{p}{1-p}$  to generate positive and negative costs. With these costs set appropriately, the optimal solution of minimum cost multicut problems not only yields an optimal cluster assignment but also estimates the number of clusters (e.g. objects to track) automatically. 

While the plain minimum cost multicut problem has shown good performance in multiple object tracking scenarios with only short range information available~\cite{tang2016multi}, the cost function actually has a rather limited expressiveness. In particular, when we want to add connectivity cues between temporally distant bounding boxes, we can only do so by inserting a direct edge into the graph. This facilitates solutions that directly connect such distant nodes even if this link is not justified by any path through space and time. 

This limitation is alleviated by the formulation of minimum cost \emph{lifted} multicuts~\cite{keupericcv}.

\subsubsection{Minimum Cost Lifted Multicut Problem}
For a given, undirected graph $G = (V, E)$ and an additional edge set $F\subseteq \binom{V}{2} \setminus E$ and any  real valued cost function $c: E \cup F \to \mathbb{R}$,
the 01 linear program written below is an instance of the
\emph{Minimum Cost Lifted Multicut Problem (LMP)}
w.r.t.~$G$, $F$ and $c$~\cite{keupericcv}:
\begin{equation}
\displaystyle\min_{y \in Y_{EF}} 
    \quad \sum_{e \in E \cup F} c_e y_e
    \label{eq:lmc}
\end{equation}
with $Y_{EF} \subseteq \{0,1\}^{E \cup F}$ the set of all $y \in \{0,1\}^{E \cup F}$ with
\begin{equation}
 \qquad\forall C \in \textnormal{cycles}(G)\ \forall e \in C:\ 
    y_e \leq \hspace{-2ex} \sum_{e' \in C \setminus \{e\}} \hspace{-2ex} y_{e'} 
\label{eq:lmc-cut1}
\end{equation}
\begin{equation}
 \forall vw \in F\ \forall P \in vw\textnormal{-paths}(G):\ 
    y_{vw} \leq \sum_{e \in P} y_e \quad
\label{eq:lmc-cut2}\end{equation}
\begin{equation}
 \hspace{-1ex} \forall vw \in F\ \forall C \in vw\textnormal{-cuts}(G):
    1 - y_{vw} \leq \sum_{e \in C} (1 - y_e) 
\label{eq:lmc-cut}
   \end{equation}

The above inequalities Eq.~\eqref{eq:lmc-cut1} make sure that, as before, the resulting edge labeling is actually inducing a decomposition of $G$. Eq.~\eqref{eq:lmc-cut2} enforces the same constraints on cycles involving edges from $F$, i.e. so called \emph{lifted} edges, and Eq.\eqref{eq:lmc-cut} makes sure that nodes that are connected via a lifted edge $e\in F$ are connected via some path along original edges  $e'\in E$ as well. Thus, this formulation allows for a generalization of the cost function to include long range information without altering the set of feasible solutions.

\paragraph{Optimization}
The minimum cost multicut problem~\eqref{eq:mc} as well as the minimum lifted multicut problem~\eqref{eq:lmc} are NP-hard \cite{bansal-2004} and even APX-hard \cite{demaine-2006,hornakova-2017}. 
Nonetheless, instances have been solved within tight bounds for example in \cite{andres-2012-globally} using a branch-and-cut approach. 
While this can be reasonably fast for some, easier problem instances, it can take arbitrarily long for others. 
Thus, primal heuristics such as the one proposed in~\cite{keupericcv} or~\cite{CGC} are often employed in practice and show convincing results in various scenarios \cite{keupericcv,tang2017multiple,insafutdinov2016deepercut}.

\subsubsection{Spatio-Temporal Tracklet Generation}
Based on this clustering formulation, detection bounding boxes can be grouped into tracklets and tracks using any available spatial or appearance information allowing to estimate pairwise cut probabilities. Since the proposed approach is unsupervised in a sense that no annotated labels are used in the training process, it is challenging to effectively learn such probabilities. To approach this challenge, we first extract reliable point matches between neighboring frames using DeepMatching~\cite{dm} as done before e.g. in \cite{tang2016multi,tang2017multiple,keuper2018motion}. Instead of learning a regression model on features derived from the resulting point matches, we simply assume that the intersection over union (IoU) of retrieved matched within pairs of detections (denoted by IoU$_{\mathrm{DM}}$) is an approximation to the true IoU. Thus, when IoU$_{\mathrm{DM}}>0.7$, we can be sure we are looking at the same object in different frames. While this rough estimation is not suitable in the actual tracking task since it clearly over-estimates the cut probability, it can be used to perform a pre-grouping of detections that definitely belong to the same person. 
The computation of pairwise cut probabilities used in the lifted multicut step for the final tracking task is described in section~\ref{subsec:affinity}.

\subsection{Deep Convolutional AutoEncoder}
\label{subsec:autoencoder}
A convolutional AutoEncoder takes an input image, compresses it into a \textit{latent space} and reconstructs it with the objective to learn meaningful features in an unsupervised manner. It consists of two parts: the encoder $f_\theta(.)$ and a decoder $g_\phi(.)$, where $\theta$ and $\phi$ are trainable parameters of the encoder and decoder, respectively. For a given input video, there are in total $n$ detections ${x_i \in X}_{i=1}^n$, the objective is to find a meaningful encoding $z_i$, where the dimension of $z_i$ is much lower than $x_i$. The used convolutional autoencoder first maps the input data into a latent space $Z$ with a non-linear function $f_\theta: X \rightarrow Z$, then decodes $Z$ to its input with $g_\phi:Z\rightarrow X$. The encoding and reconstruction is achieved by minimizing the following loss equation:

\begin{equation}
\label{eq:recloss}
\min_{\theta, \phi} \sum_{i=1}^{N} L (g(f(x_i)), x_i)
\end{equation}

where L is the least-squared loss $L(x,y)=\|x-y\|^2$. 
Similar to the work of \cite{yang2016towards}, we add an additional clustering term to minimize the distance between learned features and their cluster center $\tilde{c_i}$ from the spatio-temporal tracklet labels.

\begin{equation}
\label{eq:clusterloss}
\min_{\theta, \phi} \sum_{i=1}^{N} L (g(f(x_i)), x_i)\lambda + L (f(x_i),\tilde{c_i})(1-\lambda)
\end{equation}

The parameter $\lambda \in [0, 1]$ balances between reconstruction and clustering loss.
When choosing $0 < \lambda < 1$, the reconstruction part (Eq.~\eqref{eq:recloss}) can be considered a data-dependent regularization for the clustering.
To compute the centroid $c_i$, the whole dataset is passed through the AutoEncoder once:

\begin{equation}
\label{eq:centroid}
\tilde{c_i} = \frac{1}{N} \sum_{i=1}^{N} f(x_i)
\end{equation}

\begin{figure}[t]
	\begin{center}
		\includegraphics[width=1.0\linewidth]{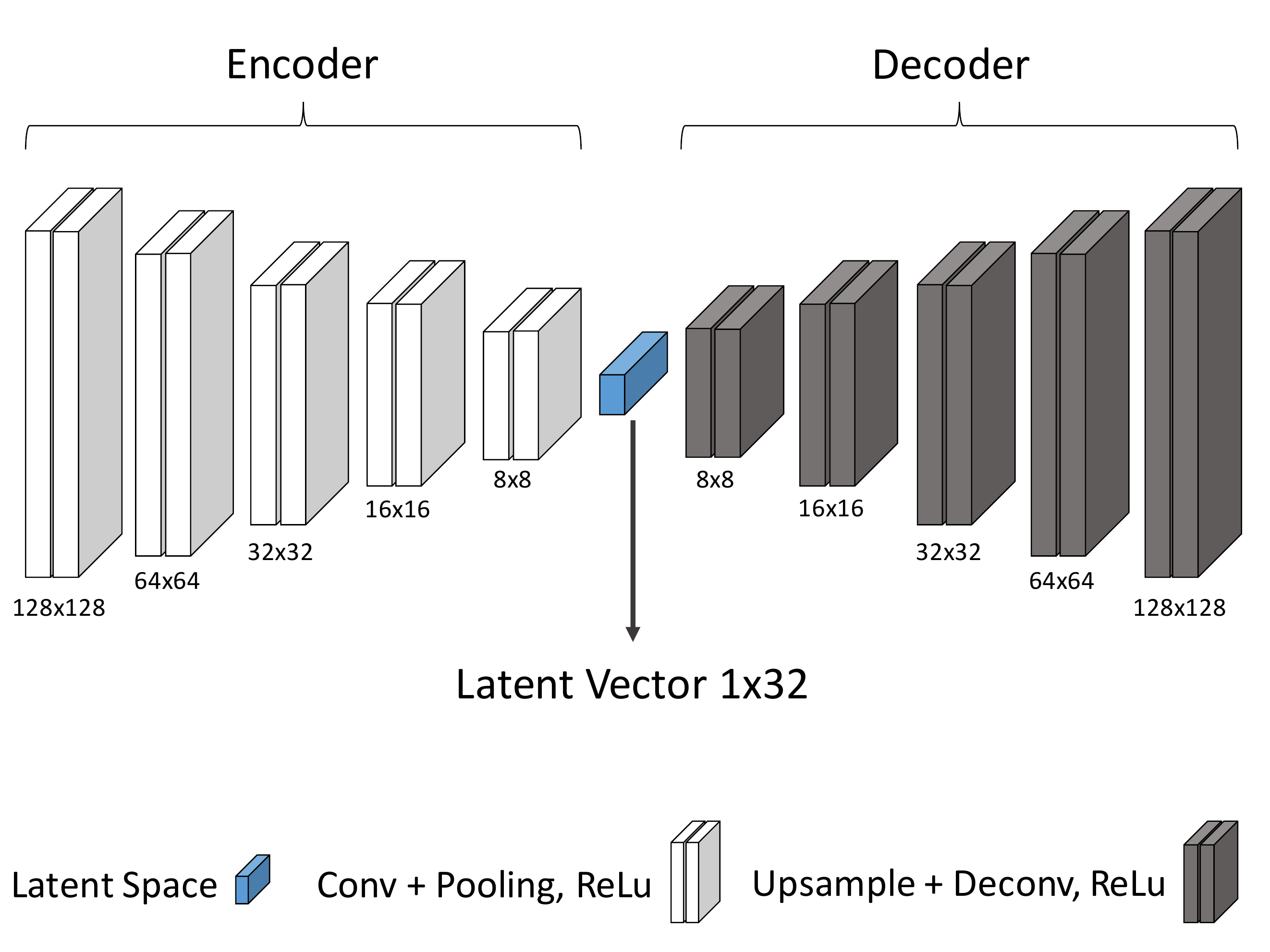}
	\end{center}
	\caption{Architecture of proposed auto-encoder. It consists of five convolutional and max-pooling layers with one fully connected at the center (Latent Space). The displayed numbers below each layer correspond to the according input size.}
	\label{fig:figure2}
\end{figure}

We use a deep AutoEncoder with five convolutional and max-pooling layers for the encoder and five up-convolutional and upsample layers for the decoder, respectively. Furthermore, batch normalization is applied on each layer and initialized using Xavier Initialization \cite{glorot2010understanding}. Inspired by U-net \cite{ronneberger2015u}, the input image size is halved after each layer while the number of filters are doubled. The size of latent space is set to 32. The input layer takes a colored image with dimension $64\times 64$ in width and height. For both the encoder and decoder, we use ReLu activation functions on each layer. Figure \ref{fig:figure2} illustrates the proposed auto-encoder architecture.

\subsection{AutoEncoder-based Affinity Measure}
\label{subsec:affinity}
We use the trained AutoEncoder to estimate the similarity of two detections $x_i$ and $x_j$ of a video sequence based on the euclidean distance in the latent space:

\begin{equation}
\label{eq:euclid}
d_{i, j} = \|f(x_i) - f(x_j)\|
\end{equation}

\begin{figure*}[t]
	\begin{center}
		\includegraphics[width=0.8\linewidth]{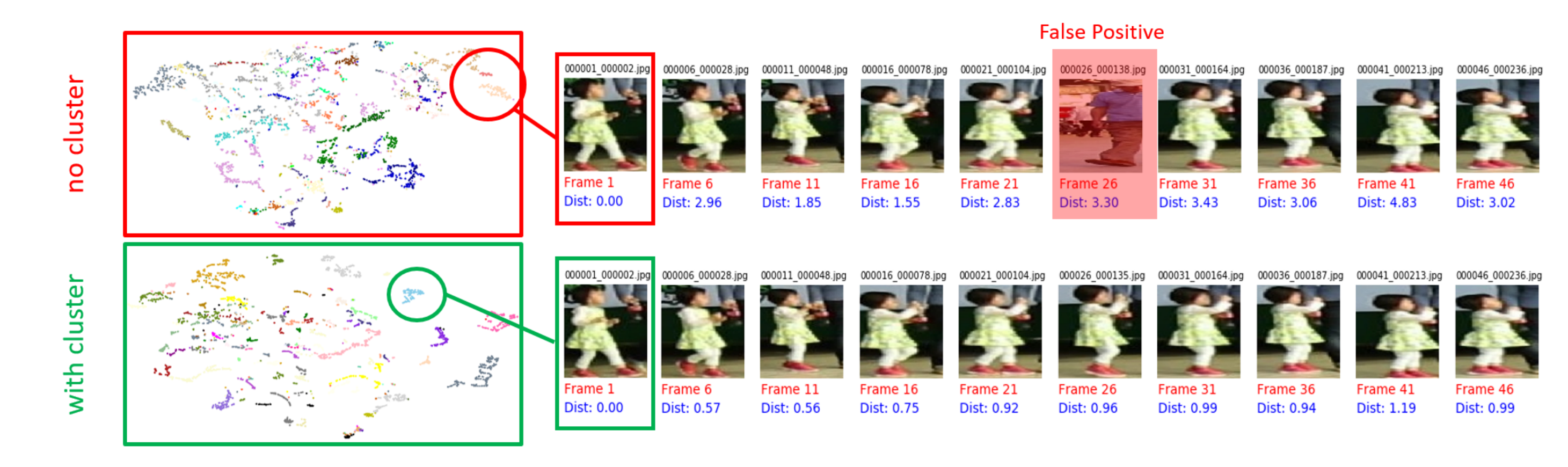}
	\end{center}
	\caption{Nearest Neighbor of the very left detection within 40 frames with a step size of 5 frames of the sequence MOT17-09-SDP. The second row shows a False Positive neighbor.}
	\label{fig:figure3}
\end{figure*}

Figure~\ref{fig:figure3} shows the nearest neighbor of a selected frame $t$ (left box marked in red) from the sequence MOT17-09 and Frame $t+n$. 
The example illustrates that the location of pair detections are close to one another in the latent space even over a long distance of up to 40 frames although false positives still appear. 
However, the example also shows that change in appearance affects the auto-encoder distance, further denoted $d_{\mathrm{AE}}$. 
For instance in the first row, frame 1 and frame 6 are very similar due to the same detection position of the person within the bounding box as well as the direction the woman is looking to (face to camera). 
At frame 41, the woman slightly turned her face to the right. 
Although the correct nearest neighbour was retrieved, the distance $d_{\mathrm{AE}}$ almost doubled.
Another observation is that the position of the bounding box influences the latent-space distance.
In the second row, the first detection from the left (frame 5), the detection of the person is slightly shifted to the left. 
At frame 15, 20 or 25, the position is slightly zoomed and $d_{\mathrm{AE}}$ increases.

\begin{figure}[!t]
	\begin{center}
		\includegraphics[width=1.0\linewidth]{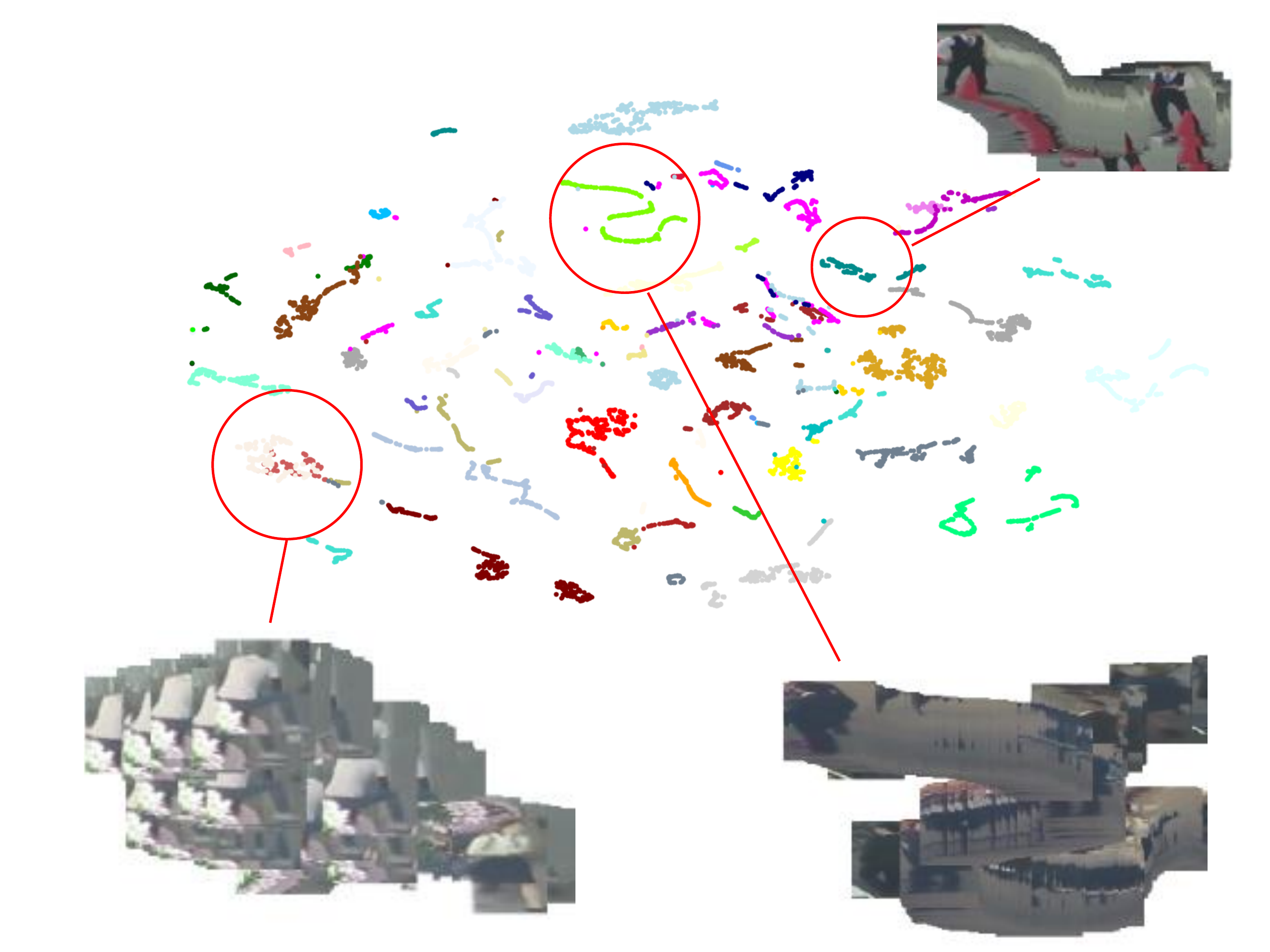}
	\end{center}
	\caption{TSNE Visualization of the latent space of the trained auto-encoder for the sequence MOT17-04 FRCNN. The colors represent the assigned person IDs.}
	\label{fig:graph4}
\end{figure}

Figure \ref{fig:graph4} shows the TSNE-Visualization \cite{maaten2008visualizing} of the latent space from the sequence MOT17-04-FRCNN. 
Our proposed auto-encoder learned the visual features without supervision. 
The different colors represent the cluster labels. 
As shown in the example circled on the bottom left, similar looking persons are very closed in the latent space: The sitting person in white shirt and the lady, wearing a white shirt.
In fact, when looking closely, the same "flower"-shaped on the bottom left corner of the sitting person is also found on the upper left corner of the white-shirted lady.
The visualization also shows that the same person may change the appearance over time (example on the bottom right). 
In the latent space, the "snake"-like shape may indicate that the viewpoint on a person may have changed over time, causing a continuous appearance change. 
When standing still, the change is minimal, which is also observed in the example on the top right corner.
While for nearby frames, we can compute pairwise cues based on the distance between latent feature representation ($d_{\mathrm{AE}}$), as well as on spatial cues (IoU$_{\mathrm{DM}}$), spatial information can not be used to associate detections over longer temporal distances. 
However, to facilitate the re-identification of objects after being fully or partly occluded, such long-range information is needed.
In these cases, we have to purely rely on the learned latent space distance $d_{\mathrm{AE}}$. 

The distance is directly casted to a binary logistic regression to compute the cut probability of the respective edge in graph $G$. The label that is used for the regression comes from the DeepMatching IoU. If IoU$_{\mathrm{DM}}(x_i,x_j)<T_{low}$ for a threshold $T_{low}$, $x_i$ and $x_j$ most certainly belong to different objects. If IoU$_{\mathrm{DM}}(x_i,x_j)>T_{high}$ for a different threshold $T_{high}$, they are very likely to match. Formally, we estimate a probability $p_e \in [0,1]$ between two detections using a feature vector $f^{(e)}$ by regressing the parameters $beta$ of a logistic function

\begin{equation} 
\label{eq:regression} p_e = \frac{1}{1+\exp(-\langle\beta, f^{(e)}\rangle)}
\end{equation} 
Thus, the costs $c_e$ can intuitively be computed by the logit. To robustly estimate these probabilities, we set $T_{low}$ and $T_{high}$ most conservatively to $0.1$ and $0.7$, respectively. 

From this partial, purely spatially induced labeling, we can estimate cut probabilities for all available features combinations, i.e. possible combinations of IoU$_\mathrm{DM}$ and $d_{\mathrm{AE}}$ within nearby frames and only $d_{\mathrm{AE}}$ for distant frames.
%
%


\section{Experiments and Results}
\label{sec:results}
We evaluated the proposed method on the MOT17 Benchmark~\cite{MOT16} for multiple person tracking. The dataset consists of 14 sequences, divided into train and test sets with 7 sequences each. For all sequences, three different detection sets are provided, from the detectors SDP\cite{yang2016exploit}, DPM\cite{felzenszwalb2010object} and FRCNN \cite{ren2015faster}, thus yielding 21 sequences in both data splits. The settings between the training and testing scenes are very similar such as moving/static camera, place of recording or view angle, such that learning-based methods usually train on the most similar training sequence for every test sequence.
For the evaluation, we use the standard CLEAR MOTA metric~\cite{MOT15}. We reported Tracking Accuracy (MOTA), Precision (MOTP), number of identity switches (IDs), mostly tracked trajectories ratio (MT) and mostly lost trajectories (ML).

\begin{table*}[t]
\small
\begin{center}
	\begin{tabular}{|c|l|c|r|r|r|r|r|r|r|}
		\hline
		No   & Features & Distance & MOTA & MOTP  & IDs & MT & ML & FP & FN \\
		\hline\hline
		1 & IoU$_{\mathrm{DM}}$ & 1-3 & 47.2 & 83.8 & 2,976 & 313 & 659 & 7,633 & 167,109 \\
		2 & $d_{\mathrm{AE}}$ & 1-3 & 35.9 & 84.2 & 4,611 & 128 & 755 & 6,726 & 204,681 \\
		3 & $d_{\mathrm{AE+C}}$ & 1-3 & 46.6 & 83.7 & 2,290 & 302 & 655 & 7,592 & 170,140 \\
		4 & IoU$_{\mathrm{DM}}+d_{\mathrm{AE}}+$IoU$_{\mathrm{DM}}*d_{\mathrm{AE}}$ & 1-3 & 49.4 & 83.5 & 1,730 & 381 & 593 & 7,536 & 161,057 \\
		5 & IoU$_{\mathrm{DM}}+d_{\mathrm{AE+C}}+$IoU$_{\mathrm{DM}}*d_{\mathrm{AE+C}}$ & 1-3 & 49.4 & 83.4 & 1,713 & 380 & 594 & 7,786 & 161,084  \\
		\hline\hline
		6 & IoU$_{\mathrm{DM}}$ & 1-5 & 47.1 & 83.6 & 2,808 & 318 & 669  & 11,015 & 164,515 \\
		7 & $d_{\mathrm{AE}}$ & 1-5 & 35.1 & 84.0 & 4,361 & 139 & 745  & 8,810 & 174,717 \\
		8 & $d_{\mathrm{AE+C}}$ & 1-5 & 44.8 & 83.6 & 2,585 & 276 & 660 & 8,810 & 174,717 \\
		9 & IoU$_{\mathrm{DM}}+d_{\mathrm{AE}}+$IoU$_{\mathrm{DM}}*d_{\mathrm{AE}}$ & 1-5 & 49.7 & 83.3 & 1,567 & 389 & 578 & 9,067 & 158,788 \\
		10 & IoU$_{\mathrm{DM}}+d_{\mathrm{AE+C}}+$IoU$_{\mathrm{DM}}*d_{\mathrm{AE+C}}$ & 1-5 & 49.8 & 83.3 & 1,569 & 388 & 580 & 8,869 & 158,715 \\
		\hline\hline
		\textbf{11} & \textbf{IoU$_{\mathrm{DM}}+d_{\mathrm{AE+C}}+$IoU$_{\mathrm{DM}}*d_{\mathrm{AE+C}}$ Lift} & \textbf{1-5} & \textbf{49.9} & \textbf{83.3} & \textbf{1,453} & \textbf{394} & \textbf{579} & \textbf{8,929} & \textbf{158,448} \\
		\hline
	\end{tabular}
\end{center}
\caption{Tracking Performance using different features on MOT17 Training Dataset. The $*$ indicates that clustering loss is applied.}
\label{tab:mcsetup}
\end{table*}
\begin{table*}
	\small
	\begin{center}
		\begin{tabular}{|l|c|c|c|c|c|c|c|c|c|}
			\hline
			Sequence & Method & \textbf{MOTA} & MOTP & IDs & IDF1 & MT & ML & FP & FN \\
			\hline\hline
			LSST17\cite{feng2019multi} & Supervised & \textbf{54.7} & 75.9 & \textbf{1,243} & \textbf{62.3} & 20.4 & 40.1 & 26,091 & \textbf{228,434}\\
			Tracktor17\cite{bergmann2019tracking} & Supervised & 53.5 & \textbf{78.0} & 2,072  & 52.3 & 19.5 & 36.6 & \textbf{12,201} & 248,047 \\
			JBNOT\cite{henschel2019multiple} & Supervised & 52.6 & 77.1 & 3,050  & 50.8 & 19.7 & 35.8 & 31,572 & 232,659 \\
			FAMNet\cite{chu2019famnet} & Supervised & 52.0 & 76.5 & 3,072  & 48.7 & 19.1 & \textbf{33.4} & 14,138 & 253,616 \\
			eTC17\cite{wang2019exploit} & Supervised & 51.9 & 76.3 & 2,288  & 58.1 & \textbf{23.1} & 35.5 & 36,164 & 232,783 \\
			\hline\hline
			Ours & Unsupervised & 48.1 & 76.7 & 2,350 & 46.0 & 17.7 & 39.9 & 16,839 & 273,819\\
			\hline
			
		\end{tabular}
	\end{center}
	\caption{Tracking performance compared to other methods on the MOT17 dataset.}
	\label{tab:mot17compare}
\end{table*}
After providing our implementation details, we report an ablation study on the training sequences of MOT17 in section~\ref{subsec:ablationstudy}. 
Our final results are discussed in section~\ref{subsec:mot17benchark}

\paragraph{Implementation Details}
\label{subsec:experiment}

Our implementation is based on the Tensorflow Deep Learning Framework. 
Our proposed convolutional AutoEncoder is used to extract features in an unsupervised manner. 
Thus no pre-training or any other ground truth is required. 
Furthermore, our pre-processing step is only limited to extracting the detections from all sequences and resizing them to the corresponding size of the AutoEncoder input layer. 
Thus the detections from the MOT17 dataset are directly fed to the auto-encoder. 
For each sequence from the dataset (MOT17-01-MOT17-14 with SDP, FRCNN and DPM), one individual auto-encoder is trained with the same setup and training parameters.
However, it is important to note that the number of detections for each individual person varies significantly. 
This is due to the fact that individual pedestrians are captured in a scene over many frames while others are quickly passing by or simply missed by the detector.
The duration of being captured by the detectors however varies. 
While some pedestrians are staying in the scene for a long time, others are passing by quickly out of the scene. 
This also results different cluster size at the end.
To balance this, randomized batches of detections are applied during the training, where each batch contains only images from one single frame. 
This way, one iteration of training contains only detections from unique persons. The initial learning rate is set to $\alpha=0.001$ and decays exponentially by a factor of 10 over time. The balancing parameter between reconstruction and clustering loss is set to $\lambda = 0$ at the beginning in order to first learn the visual features of the video sequences. 
Later on, the cluster information is included in the training. 
The number of training epochs varies for different sequences and detectors. 
Once the visual features have been learned, $\lambda$ is set to $0.95$ to encode the appearance variations from the spatio-temporal clusters into the latent space of the auto-encoder.

To transform detection clusters into actual tracks, we follow the procedure proposed in~\cite{tang2016multi}, i.e. from all detection within one cluster, we select per frame the one with the best detection score. Clusters containing less than 5 detections are completely removed and gaps in the resulting tracklets are filled using bilinear interpolation. 

\subsection{Ablation Study}
\label{subsec:ablationstudy}

We investigated feature setups in the minimum cost multicut framework. 
The cut probability between pairs of nodes are computed using a logistic regression function. 
Adding new features directly affects the edge cost between pairs thus resulting in different clustering performances. 
Here, we investigate the extent to which our proposed appearance model improves the tracking performance.
All studies were conducted on the complete MOT17 training dataset. 

Table \ref{tab:mcsetup} shows the evaluated setups and the resulting tracking performance scores. 
In total, we conducted 11 experiments. 
The column \textit{Features} lists the added features to the logistic regression model. The temporal distances over which bounding boxes are connected in the graph are marked in the column \textit{Distances}. 
The tracking accuracy of experiment 1 and 6, which uses IoU$_{\mathrm{DM}}$ only, is 47.2\% and 47.1\%, respectively. 
Experiment 2+3 (7+8) compare the different auto-encoder models: the euclidean distance ($d_{\mathrm{AE}}$) from the auto-encoder latent space is computed in order to estimate the similarity of each pair detections. 
Here, $d_{\mathrm{AE}}$ denotes the latent space distance before adding the clustering loss, $d_{\mathrm{AE+C}}$ denotes the distance after training of the auto-encoder with the clusterig loss, i.e. our proposed appearance method.

\begin{table*}[t]
	\small
	\begin{center}
		\begin{tabular}{|l|r|r|r|r|r|r|r|r|r|r|r|r|}
			\hline
			Sequence & \textbf{MOTA} & MOTP& MOTAL& Rcll & Prcn & IDs & F1 & MT & ML & FP & FN \\
			\hline\hline
			MOT17-01-FRCNN & \textbf{28.3} & 76.6 & 28.8 & 48.8 & 71 & 38 & 32.8 & 6 & 10 & 1285 & 3304\\
			MOT17-03-FRCNN & \textbf{57.3} & 77.9 & 57.6 & 59.1 & 97.5 & 299 & 48.8 & 36 & 28 & 1617 & 42773\\
			MOT17-06-FRCNN & \textbf{55.0} & 78.7 & 55.8 & 59 & 94.9 & 96 & 58.7 & 50 & 72 & 370 & 4832\\
			MOT17-07-FRCNN & \textbf{34.7} & 75.1 & 35.4 & 41.8 & 86.7 & 124 & 38.7 & 5 & 18 & 1080 & 9834\\
			MOT17-08-FRCNN & \textbf{23.2} & 79.6 & 23.5 & 26.5 & 90 & 75 & 29.1 & 7 & 37 & 622 & 15531\\
			MOT17-12-FRCNN & \textbf{39.2} & 78.4 & 39.5 & 43.9 & 91 & 26 & 51.6 & 12 & 45 & 378 & 4863\\
			MOT17-14-FRCNN & \textbf{23.1} & 71.6 & 23.9 & 35 & 76 & 161 & 36.2 & 12 & 78 & 2041 & 12015\\
			MOT17-01-SDP & \textbf{42.8} & 73.8 & 43.7 & 58.9 & 79.5 & 63 & 38.4 & 8 & 4 & 979 & 2649\\
			MOT17-03-SDP & \textbf{73.3} & 77.6 & 73.6 & 74.8 & 98.4 & 356 & 60.2 & 69 & 15 & 1241 & 26357\\
			MOT17-06-SDP & \textbf{57.7} & 76.5 & 58.6 & 63.2 & 93.3 & 104 & 59.6 & 74 & 73 & 537 & 4340\\
			MOT17-07-SDP & \textbf{48.2} & 75.3 & 48.9 & 54.5 & 90.7 & 118 & 43.7 & 16 & 15 & 941 & 7689\\
			MOT17-08-SDP & \textbf{29.6} & 78.2 & 30.3 & 32.8 & 93.1 & 144 & 31.1 & 10 & 35 & 516 & 14202\\
			MOT17-12-SDP & \textbf{42.3} & 78.5 & 42.6 & 51.2 & 85.8 & 34 & 54.9 & 19 & 39 & 737 & 4232\\
			MOT17-14-SDP & \textbf{34.5} & 72.1 & 35.3 & 45.6 & 81.7 & 159 & 41.6 & 12 & 54 & 1892 & 10061\\
			MOT17-01-DPM & \textbf{26.7} & 72.5 & 27.0 & 28.5 & 95.1 & 26 & 27.2 & 3 & 12 & 94 & 4611\\
			MOT17-03-DPM & \textbf{46.2} & 75.7 & 46.4 & 47.8 & 97.2 & 272 & 39.5 & 21 & 41 & 1434 & 54661\\
			MOT17-06-DPM & \textbf{44.4} & 73.8 & 45.0 & 46.4 & 97 & 66 & 51.3 & 26 & 120 & 169 & 6313\\
			MOT17-07-DPM & \textbf{35.2} & 74.1 & 35.5 & 37.3 & 95.4 & 62 & 38.4 & 5 & 30 & 305 & 10584\\
			MOT17-08-DPM & \textbf{25.1} & 78.9 & 25.4 & 26.5 & 95.9 & 71 & 28.8 & 6 & 39 & 240 & 15521\\
			MOT17-12-DPM & \textbf{40.9} & 76.7 & 41.1 & 43.9 & 94 & 21 & 50.6 & 13 & 50 & 241 & 4860\\
			MOT17-14-DPM & \textbf{20.2} & 75.0 & 20.4 & 21.1 & 97 & 35 & 27.6 & 8 & 116 & 120 & 14587\\

			\hline\hline
			\textbf{Total} & \textbf{48.1} & 76.7& 48.5 & 51.5 & 94.5 & 2,350 & 46.0 & 7.3 & 32.9 & 16,839 & 273,819\\
			\hline
			
		\end{tabular}
	\end{center}
	\caption{Tracking Performance on Test Dataset of MOT17.}
	\label{tab:mot17test}
\end{table*}

The benefit of using the clustering loss on the model training is obvious: for both distances (1-3 and 1-5), the performance is significantly higher. 
For distance 1-3, $d_{\mathrm{AE+C}}$ has a tracking accuracy of 46.6 compared to $d_{\mathrm{AE}}$ (35.9) and for distance 1-5, the MOTA scores are 44.8 and 35.1 for $d_{\mathrm{AE+C}}$ and $d_{\mathrm{AE}}$, respectively. 
Although the scores are lower than using IoU$_{\mathrm{DM}}$, combining them both together increases the performance further. 
This is shown in experiment 4+5 and 9+10, where the best score is achieved with in experiment 10 (proposed method).
We also observe that the number of identity switches (IDs) is reduced with our setup.
Finally, we add a long range edges and solve solve minimum cost lifted multicut problems on $G$. 
The lifted edges are inserted in a sparsely for 10, 20 and 30 frames distances. 
Our best performance is achieved using the setup of experiment 11 with a MOTA of 49.9\% using all model components.

\subsection{Results}
\label{subsec:mot17benchark}

Here, we present and discuss our final tracking results on the MOT17 test dataset. 
An overview of all scenes and detectors is shown in table \ref{tab:mot17test}. 
Compared to the performance on the training dataset, the MOTA score of our unsupervised approach is slightly lower (Training: 49.9\% vs. Testing: 48.1\%) which can be due to a slight dataset drift. 
However, we consider a tracking accuracy difference of 1.8\% as still acceptable since no supervision on any model is applied nor excessive parameter tuning was conducted, over-fitting of data is unlikely. 
The only parameter was choosing the lifted edge range, which increased the tracking accuracy on the training data by roughly 0.1\% and reduced the number of IDs by 116.
The best performance is achieved in conjunction with the SDP-detector while the performance on the noisier DPM detections is weaker. 
Scene MOT17-08 performs the poorest over all three detectors (the MOTA score of all three detectors are below 30\%).

\newpage
We compare our method with the current top five best reported tracking methods LSST17\cite{feng2019multi}, Tracktor17\cite{bergmann2019tracking}, JBNOT\cite{henschel2019multiple}, FAMNet\cite{chu2019famnet} and eTC17\cite{wang2019exploit}.
We consider a tracking method as supervised when ground truth data is used (for example label data for learning a regression function) or if any pre-trained model.
Table \ref{tab:mot17compare} gives an overview of the scores in different metrics that is being evaluated. The best on each category is marked in bold. 
Our proposed method is competitive given the fact that no pre-trained model or any other ground truth is employed. 

Without using any of the provided human annotations and with identical parameter settings for all sequences and detectors, our resulting MOTA scores are very close to the state-o-the-art models.

\section{Conclusion}
\label{sec:conclusion}

In this work, we presented an approach towards tracking of multiple persons without the supervision from human annotations. Combining the visual features learned from an auto-encoder with spatio-temporal cues, we are able to automatically create robust appearance cues enabling multiple person tracking over a long distance. The result of our proposed method achieves a tracking accuracy of 48.1\% on the MOT17 benchmark. 
To the best of our knowledge, we are the first to propose a fully unsupervised but competitive approach to pedestrian tracking on the MOT benchmarks. However, in light of the very diverse scenarios in which pedestrian tracking is to be used in practice, the need for such unsupervised tracking methods is obvious.

{\small
\bibliographystyle{IEEEtran}
\bibliography{egbib, paper}

\begin{thebibliography}{10}
\providecommand{\url}[1]{#1}
\csname url@samestyle\endcsname
\providecommand{\newblock}{\relax}
\providecommand{\bibinfo}[2]{#2}
\providecommand{\BIBentrySTDinterwordspacing}{\spaceskip=0pt\relax}
\providecommand{\BIBentryALTinterwordstretchfactor}{4}
\providecommand{\BIBentryALTinterwordspacing}{\spaceskip=\fontdimen2\font plus
\BIBentryALTinterwordstretchfactor\fontdimen3\font minus
  \fontdimen4\font\relax}
\providecommand{\BIBforeignlanguage}[2]{{%
\expandafter\ifx\csname l@#1\endcsname\relax
\typeout{** WARNING: IEEEtran.bst: No hyphenation pattern has been}%
\typeout{** loaded for the language `#1'. Using the pattern for}%
\typeout{** the default language instead.}%
\else
\language=\csname l@#1\endcsname
\fi
#2}}
\providecommand{\BIBdecl}{\relax}
\BIBdecl

\bibitem{zamir2012gmcp}
A.~R. Zamir, A.~Dehghan, and M.~Shah, ``Gmcp-tracker: Global multi-object
  tracking using generalized minimum clique graphs,'' in \emph{Computer
  Vision--ECCV 2012}.\hskip 1em plus 0.5em minus 0.4em\relax Springer, 2012,
  pp. 343--356.

\bibitem{henschel2017improvements}
R.~Henschel, L.~Leal-Taix{\'e}, D.~Cremers, and B.~Rosenhahn, ``Improvements to
  frank-wolfe optimization for multi-detector multi-object tracking,''
  \emph{arXiv preprint arXiv:1705.08314}, 2017.

\bibitem{tang2016multi}
S.~Tang, B.~Andres, M.~Andriluka, and B.~Schiele, ``Multi-person tracking by
  multicut and deep matching,'' in \emph{European Conference on Computer
  Vision}.\hskip 1em plus 0.5em minus 0.4em\relax Springer, 2016, pp. 100--111.

\bibitem{tang2017multiple}
S.~Tang, M.~Andriluka, B.~Andres, and B.~Schiele, ``Multiple people tracking by
  lifted multicut and person reidentification,'' in \emph{Proceedings of the
  IEEE Conference on Computer Vision and Pattern Recognition}, 2017, pp.
  3539--3548.

\bibitem{luo2014multiple}
W.~Luo, J.~Xing, A.~Milan, X.~Zhang, W.~Liu, X.~Zhao, and T.-K. Kim, ``Multiple
  object tracking: A literature review,'' \emph{arXiv preprint
  arXiv:1409.7618}, 2014.

\bibitem{MOT16}
\BIBentryALTinterwordspacing
A.~Milan, L.~Leal-Taix\'{e}, I.~Reid, S.~Roth, and K.~Schindler, ``{MOT}16: {A}
  benchmark for multi-object tracking,'' \emph{arXiv:1603.00831 [cs]}, Mar.
  2016, arXiv: 1603.00831. [Online]. Available:
  \url{http://arxiv.org/abs/1603.00831}
\BIBentrySTDinterwordspacing

\bibitem{yoon2018online}
Y.-c. Yoon, A.~Boragule, Y.-m. Song, K.~Yoon, and M.~Jeon, ``Online
  multi-object tracking with historical appearance matching and scene adaptive
  detection filtering,'' in \emph{2018 15th IEEE International Conference on
  Advanced Video and Signal Based Surveillance (AVSS)}.\hskip 1em plus 0.5em
  minus 0.4em\relax IEEE, 2018, pp. 1--6.

\bibitem{feng2019multi}
W.~Feng, Z.~Hu, W.~Wu, J.~Yan, and W.~Ouyang, ``Multi-object tracking with
  multiple cues and switcher-aware classification,'' \emph{arXiv preprint
  arXiv:1901.06129}, 2019.

\bibitem{Pirsiavash:2011:GOG}
H.~Pirsiavash, D.~Ramanan, and C.~C. Fowlkes, ``Globally-optimal greedy
  algorithms for tracking a variable number of objects,'' in \emph{CVPR}, 2011.

\bibitem{Andriyenko2012CVPR}
A.~Andriyenko, K.~Schindler, and S.~Roth, ``Discrete-continuous optimization
  for multi-target tracking,'' in \emph{CVPR}, 2012.

\bibitem{Huang:2008:ROT}
C.~Huang, B.~Wu, and R.~Nevatia, ``Robust object tracking by hierarchical
  association of detection responses,'' in \emph{ECCV}, 2008.

\bibitem{AndrilukaCVPR2010}
M.~Andriluka, S.~Roth, and B.~Schiele, ``Monocular 3d pose estimation and
  tracking by detection,'' in \emph{CVPR}, June 2010.

\bibitem{FragkiadakiECCV12}
K.~Fragkiadaki, W.~Zhang, G.~Zhang, and J.~Shi, ``Two-granularity tracking:
  Mediating trajectory and detection graphs for tracking under occlusions,'' in
  \emph{ECCV}, 2012.

\bibitem{Zamir:2012:GMC}
A.~R. Zamir, A.~Dehghan, and M.~Shah, ``{GMCP}-{T}racker: Global multi-object
  tracking using generalized minimum clique graphs,'' in \emph{ECCV}, 2012.

\bibitem{Henschel:2014:EMP}
R.~Henschel, L.~Leal-Taixe, and B.~Rosenhahn, ``Efficient multiple people
  tracking using minimum cost arborescences,'' in \emph{GCPR}, 2014.

\bibitem{tang14ijcv}
S.~Tang, M.~Andriluka, and B.~Schiele, ``Detection and tracking of occluded
  people,'' \emph{IJCV}, 2014.

\bibitem{DBLP:journals/corr/HenschelLCR17}
\BIBentryALTinterwordspacing
R.~Henschel, L.~Leal{-}Taix{\'{e}}, D.~Cremers, and B.~Rosenhahn,
  ``Improvements to frank-wolfe optimization for multi-detector multi-object
  tracking,'' \emph{CoRR}, vol. abs/1705.08314, 2017. [Online]. Available:
  \url{http://arxiv.org/abs/1705.08314}
\BIBentrySTDinterwordspacing

\bibitem{Shitrit:2011:TMP}
H.~B. Shitrit, J.~Berclaz, F.~Fleuret, and P.~Fua, ``Tracking multiple people
  under global appearance constraints,'' in \emph{ICCV}, 2011.

\bibitem{wang-et-al-2014}
X.~Wang, E.~Turetken, F.~Fleuret, and P.~Fua, ``Tracking interacting objects
  optimally using integer programming,'' in \emph{ECCV}, 2014.

\bibitem{10.1007/978-3-319-16817-3_29}
R.~Kumar, G.~Charpiat, and M.~Thonnat, ``Multiple object tracking by efficient
  graph partitioning,'' in \emph{Computer Vision -- ACCV 2014}, D.~Cremers,
  I.~Reid, H.~Saito, and M.-H. Yang, Eds.\hskip 1em plus 0.5em minus
  0.4em\relax Cham: Springer International Publishing, 2015, pp. 445--460.

\bibitem{7503631}
Y.~T. Tesfaye, E.~Zemene, M.~Pelillo, and A.~Prati, ``Multi-object tracking
  using dominant sets,'' \emph{IET Computer Vision}, vol.~10, no.~4, pp.
  289--297, 2016.

\bibitem{WojekECCV10}
C.~Wojek, S.~Roth, K.~Schindler, and B.~Schiele, ``Monocular 3d scene modeling
  and inference: Understanding multi-object traffic scenes,'' in \emph{ECCV},
  2010.

\bibitem{WojekPAMI2013}
C.~Wojek, S.~Walk, S.~Roth, K.~Schindler, and B.~Schiele, ``Monocular visual
  scene understanding: Understanding multi-object traffic scenes,'' \emph{IEEE
  TPAMI}, 2013.

\bibitem{networkflow1}
H.~Pirsiavash, D.~Ramanan, and C.~C. Fowlkes, ``Globally-optimal greedy
  algorithms for tracking a variable number of objects,'' in \emph{CVPR}, 2011.

\bibitem{networkflow2}
X.~Wang, E.~Turetken, F.~Fleuret, and P.~Fua, ``Tracking interacting objects
  optimally using integer programming,'' in \emph{ECCV}, 2014.

\bibitem{Chari2015OnPC}
V.~Chari, S.~Lacoste-Julien, I.~Laptev, and J.~Sivic, ``On pairwise costs for
  network flow multi-object tracking,'' \emph{2015 IEEE Conference on Computer
  Vision and Pattern Recognition (CVPR)}, pp. 5537--5545, 2015.

\bibitem{keuper2018motion}
M.~Keuper, S.~Tang, B.~Andres, T.~Brox, and B.~Schiele, ``Motion segmentation
  \& multiple object tracking by correlation co-clustering,'' \emph{IEEE
  transactions on pattern analysis and machine intelligence}, 2018.

\bibitem{henschel2018fusion}
R.~Henschel, L.~Leal-Taix{\'e}, D.~Cremers, and B.~Rosenhahn, ``Fusion of head
  and full-body detectors for multi-object tracking,'' in \emph{Computer Vision
  and Pattern Recognition Workshops (CVPRW)}, 2018.

\bibitem{henschel2019multiple}
R.~Henschel, Y.~Zou, and B.~Rosenhahn, ``Multiple people tracking using body
  and joint detections,'' in \emph{Proceedings of the IEEE Conference on
  Computer Vision and Pattern Recognition Workshops}, 2019, pp. 0--0.

\bibitem{keuper2015efficient}
M.~Keuper, E.~Levinkov, N.~Bonneel, G.~Lavou{\'e}, T.~Brox, and B.~Andres,
  ``Efficient decomposition of image and mesh graphs by lifted multicuts,'' in
  \emph{Proceedings of the IEEE International Conference on Computer Vision},
  2015, pp. 1751--1759.

\bibitem{keuper2016multi}
M.~Keuper, S.~Tang, Y.~Zhongjie, B.~Andres, T.~Brox, and B.~Schiele, ``A
  multi-cut formulation for joint segmentation and tracking of multiple
  objects,'' \emph{arXiv preprint arXiv:1607.06317}, 2016.

\bibitem{kumar2014multiple}
R.~Kumar, G.~Charpiat, and M.~Thonnat, ``Multiple object tracking by efficient
  graph partitioning,'' in \emph{Asian Conference on Computer Vision}.\hskip
  1em plus 0.5em minus 0.4em\relax Springer, 2014, pp. 445--460.

\bibitem{ma2018trajectory}
C.~Ma, C.~Yang, F.~Yang, Y.~Zhuang, Z.~Zhang, H.~Jia, and X.~Xie, ``Trajectory
  factory: Tracklet cleaving and re-connection by deep siamese bi-gru for
  multiple object tracking,'' \emph{arXiv preprint arXiv:1804.04555}, 2018.

\bibitem{bergmann2019tracking}
P.~Bergmann, T.~Meinhardt, and L.~Leal-Taixe, ``Tracking without bells and
  whistles,'' \emph{arXiv preprint arXiv:1903.05625}, 2019.

\bibitem{kim2015multiple}
C.~Kim, F.~Li, A.~Ciptadi, and J.~M. Rehg, ``Multiple hypothesis tracking
  revisited,'' in \emph{Proceedings of the IEEE International Conference on
  Computer Vision}, 2015, pp. 4696--4704.

\bibitem{8533372}
H.~Sheng, J.~Chen, Y.~Zhang, W.~Ke, Z.~Xiong, and J.~Yu, ``Iterative multiple
  hypothesis tracking with tracklet-level association,'' \emph{IEEE
  Transactions on Circuits and Systems for Video Technology}, pp. 1--1, 2018.

\bibitem{chen2017enhancing}
J.~Chen, H.~Sheng, Y.~Zhang, and Z.~Xiong, ``Enhancing detection model for
  multiple hypothesis tracking,'' in \emph{Conf. on Computer Vision and Pattern
  Recognition Workshops}, 2017, pp. 2143--2152.

\bibitem{li2018unsupervised}
M.~Li, X.~Zhu, and S.~Gong, ``Unsupervised person re-identification by deep
  learning tracklet association,'' in \emph{Proceedings of the European
  Conference on Computer Vision (ECCV)}, 2018, pp. 737--753.

\bibitem{lv2018unsupervised}
J.~Lv, W.~Chen, Q.~Li, and C.~Yang, ``Unsupervised cross-dataset person
  re-identification by transfer learning of spatial-temporal patterns,'' in
  \emph{Proceedings of the IEEE Conference on Computer Vision and Pattern
  Recognition}, 2018, pp. 7948--7956.

\bibitem{demaine-2006}
E.~D. Demaine, D.~Emanuel, A.~Fiat, and N.~Immorlica, ``Correlation clustering
  in general weighted graphs,'' \emph{Theoretical Computer Science}, vol. 361,
  no. 2--3, pp. 172--187, 2006.

\bibitem{keupericcv}
M.~Keuper, E.~Levinkov, N.~Bonneel, G.~Lavoue, T.~Brox, and B.~Andres,
  ``Efficient decomposition of image and mesh graphs by lifted multicuts,'' in
  \emph{ICCV}, 2015.

\bibitem{chopra-1993}
S.~Chopra and M.~Rao, ``The partition problem,'' \emph{Mathematical
  Programming}, vol.~59, no. 1--3, pp. 87--115, 1993.

\bibitem{bansal-2004}
N.~Bansal, A.~Blum, and S.~Chawla, ``Correlation clustering,'' \emph{Machine
  Learning}, vol.~56, no. 1--3, pp. 89--113, 2004.

\bibitem{hornakova-2017}
A.~Hor\v{n}\'akov\'a, J.-H. Lange, and B.~Andres, ``Analysis and optimization
  of graph decompositions by lifted multicuts,'' in \emph{ICML}, 2017.

\bibitem{andres-2012-globally}
B.~Andres, T.~Kr{\"o}ger, K.~L. Briggman, W.~Denk, N.~Korogod, G.~Knott,
  U.~K{\"o}the, and F.~A. Hamprecht, ``Globally optimal closed-surface
  segmentation for connectomics,'' in \emph{ECCV}, 2012.

\bibitem{CGC}
T.~Beier, T.~Kroeger, J.~Kappes, U.~Kothe, and F.~Hamprecht, ``Cut, glue, \&
  cut: A fast, approximate solver for multicut partitioning,'' in \emph{CVPR},
  2014.

\bibitem{insafutdinov2016deepercut}
\BIBentryALTinterwordspacing
E.~Insafutdinov, L.~Pishchulin, B.~Andres, M.~Andriluka, and B.~Schieke,
  ``Deepercut: A deeper, stronger, and faster multi-person pose estimation
  model,'' in \emph{European Conference on Computer Vision (ECCV)}, 2016.
  [Online]. Available: \url{http://arxiv.org/abs/1605.03170}
\BIBentrySTDinterwordspacing

\bibitem{dm}
\BIBentryALTinterwordspacing
J.~Revaud, P.~Weinzaepfel, Z.~Harchaoui, and C.~Schmid, ``Deep convolutional
  matching,'' \emph{CoRR}, vol. abs/1506.07656, 2015. [Online]. Available:
  \url{http://arxiv.org/abs/1506.07656}
\BIBentrySTDinterwordspacing

\bibitem{yang2016towards}
B.~Yang, X.~Fu, N.~D. Sidiropoulos, and M.~Hong, ``Towards k-means-friendly
  spaces: Simultaneous deep learning and clustering,'' \emph{arXiv preprint
  arXiv:1610.04794}, 2016.

\bibitem{glorot2010understanding}
X.~Glorot and Y.~Bengio, ``Understanding the difficulty of training deep
  feedforward neural networks,'' in \emph{Proceedings of the thirteenth
  international conference on artificial intelligence and statistics}, 2010,
  pp. 249--256.

\bibitem{ronneberger2015u}
O.~Ronneberger, P.~Fischer, and T.~Brox, ``U-net: Convolutional networks for
  biomedical image segmentation,'' in \emph{International Conference on Medical
  image computing and computer-assisted intervention}.\hskip 1em plus 0.5em
  minus 0.4em\relax Springer, 2015, pp. 234--241.

\bibitem{maaten2008visualizing}
L.~v.~d. Maaten and G.~Hinton, ``Visualizing data using t-sne,'' \emph{Journal
  of machine learning research}, vol.~9, no. Nov, pp. 2579--2605, 2008.

\bibitem{yang2016exploit}
F.~Yang, W.~Choi, and Y.~Lin, ``Exploit all the layers: Fast and accurate cnn
  object detector with scale dependent pooling and cascaded rejection
  classifiers,'' in \emph{Proceedings of the IEEE conference on computer vision
  and pattern recognition}, 2016, pp. 2129--2137.

\bibitem{felzenszwalb2010object}
P.~F. Felzenszwalb, R.~B. Girshick, D.~McAllester, and D.~Ramanan, ``Object
  detection with discriminatively trained part-based models,'' \emph{IEEE
  transactions on pattern analysis and machine intelligence}, vol.~32, no.~9,
  pp. 1627--1645, 2010.

\bibitem{ren2015faster}
S.~Ren, K.~He, R.~Girshick, and J.~Sun, ``Faster r-cnn: Towards real-time
  object detection with region proposal networks,'' in \emph{Advances in neural
  information processing systems}, 2015, pp. 91--99.

\bibitem{MOT15}
L.~Leal-Taixé, A.~Milan, I.~Reid, S.~Roth, and K.~Schindler, ``Motchallenge
  2015: Towards a benchmark for multi-target tracking,''
  \emph{arXiv:1504.01942}, 2015.

\bibitem{chu2019famnet}
P.~Chu and H.~Ling, ``Famnet: Joint learning of feature, affinity and
  multi-dimensional assignment for online multiple object tracking,'' in
  \emph{Proceedings of the IEEE International Conference on Computer Vision},
  2019, pp. 6172--6181.

\bibitem{wang2019exploit}
G.~Wang, Y.~Wang, H.~Zhang, R.~Gu, and J.-N. Hwang, ``Exploit the connectivity:
  Multi-object tracking with trackletnet,'' in \emph{Proceedings of the 27th
  ACM International Conference on Multimedia}, 2019, pp. 482--490.

\end{thebibliography}
}

\end{document}